\documentclass[10pt,twocolumn,letterpaper]{article}

\usepackage{iccv}

\makeatletter
\@namedef{ver@everyshi.sty}{}
\makeatother

\usepackage[dvipsnames,svgnames,x11names]{xcolor}
\usepackage{tikzextern}
\usepackage{pgffor}

\newcommand{\extfig}[2]{\tikzsetnextfilename{fig/extern/#1}{#2}}
\newcommand{\noextfig}[1]{!!!}
\newcommand{\extdata}[1]{}

\makeatletter
\renewcommand\paragraph{\@startsection{paragraph}{4}{\z@}{1ex}{-1em}{\normalfont\normalsize\bfseries}}
\makeatother

\usepackage{times}
\usepackage{epsfig}
\usepackage{graphicx}
\usepackage{amsmath}
\usepackage{amssymb}

\usepackage{times}
\usepackage{epsfig}
\usepackage{graphicx}
\usepackage{amsmath}
\usepackage{amssymb}

\usepackage{times}
\usepackage{epsfig}
\usepackage{graphicx}
\usepackage{amsmath}
\usepackage{amssymb}

\usepackage[utf8]{inputenc} 
\usepackage[T1]{fontenc}    
\usepackage{url}            
\usepackage{booktabs}       
\usepackage{amsfonts}       
\usepackage{nicefrac}       
\usepackage{bbm}            
\usepackage{enumitem}
\usepackage[accsupp]{axessibility}  
\usepackage{float}
\usepackage{microtype}
\usepackage[ruled,vlined,linesnumbered]{algorithm2e}

\usepackage[breaklinks=true,colorlinks,bookmarks=false]{hyperref}

\iccvfinalcopy 

\def\iccvPaperID{****} 

\ificcvfinal\fi

\begin{document}

\title{Iterative label cleaning for transductive and semi-supervised few-shot learning}

\author{
Michalis Lazarou$^1$ \ \ \ \ Tania Stathaki$^1$ \ \ \ \ Yannis Avrithis$^2$\\
$^1$Imperial College London\\
$^2$Inria, Univ Rennes, CNRS, IRISA\\
}


\newcommand{\head}[1]{{\smallskip\noindent\textbf{#1}}}
\newcommand{\alert}[1]{{\color{red}{#1}}}
\newcommand{\sm}{\scriptsize}
\newcommand{\eq}[1]{(\ref{eq:#1})}

\newcommand{\Th}[1]{\textsc{#1}}
\newcommand{\mr}[2]{\multirow{#1}{*}{#2}}
\newcommand{\mc}[2]{\multicolumn{#1}{c}{#2}}
\newcommand{\tb}[1]{\textbf{#1}}
\newcommand{\ch}{\checkmark}

\newcommand{\red}[1]{{\color{red}{#1}}}
\newcommand{\blue}[1]{{\color{blue}{#1}}}
\newcommand{\green}[1]{{\color{green}{#1}}}
\newcommand{\gray}[1]{{\color{gray}{#1}}}

\newcommand{\citeme}[1]{\red{[XX]}}
\newcommand{\refme}[1]{\red{(XX)}}

\newcommand{\fig}[2][1]{\includegraphics[width=#1\columnwidth]{fig/#2}}
\newcommand{\figh}[2][1]{\includegraphics[height=#1\columnwidth]{fig/#2}}


\newcommand{\tran}{^\top}
\newcommand{\mtran}{^{-\top}}
\newcommand{\zcol}{\mathbf{0}}
\newcommand{\zrow}{\zcol\tran}

\newcommand{\ind}{\mathbbm{1}}
\newcommand{\expect}{\mathbb{E}}
\newcommand{\nat}{\mathbb{N}}
\newcommand{\zahl}{\mathbb{Z}}
\newcommand{\real}{\mathbb{R}}
\newcommand{\proj}{\mathbb{P}}
\newcommand{\prob}{\mathbf{Pr}}
\newcommand{\normal}{\mathcal{N}}

\newcommand{\mif}{\textrm{if}\ }
\newcommand{\other}{\textrm{otherwise}}
\newcommand{\minimize}{\textrm{minimize}\ }
\newcommand{\maximize}{\textrm{maximize}\ }
\newcommand{\st}{\textrm{subject\ to}\ }

\newcommand{\id}{\operatorname{id}}
\newcommand{\const}{\operatorname{const}}
\newcommand{\sgn}{\operatorname{sgn}}
\newcommand{\var}{\operatorname{Var}}
\newcommand{\mean}{\operatorname{mean}}
\newcommand{\trace}{\operatorname{tr}}
\newcommand{\diag}{\operatorname{diag}}
\newcommand{\vect}{\operatorname{vec}}
\newcommand{\cov}{\operatorname{cov}}
\newcommand{\sign}{\operatorname{sign}}
\newcommand{\prj}{\operatorname{proj}}

\newcommand{\softmax}{\operatorname{softmax}}
\newcommand{\clip}{\operatorname{clip}}

\newcommand{\defn}{\mathrel{:=}}
\newcommand{\peq}{\mathrel{+\!=}}
\newcommand{\meq}{\mathrel{-\!=}}

\newcommand{\floor}[1]{\left\lfloor{#1}\right\rfloor}
\newcommand{\ceil}[1]{\left\lceil{#1}\right\rceil}
\newcommand{\inner}[1]{\left\langle{#1}\right\rangle}
\newcommand{\norm}[1]{\left\|{#1}\right\|}
\newcommand{\abs}[1]{\left|{#1}\right|}
\newcommand{\frob}[1]{\norm{#1}_F}
\newcommand{\card}[1]{\left|{#1}\right|\xspace}
\newcommand{\diff}{\mathrm{d}}
\newcommand{\der}[3][]{\frac{d^{#1}#2}{d#3^{#1}}}
\newcommand{\pder}[3][]{\frac{\partial^{#1}{#2}}{\partial{#3^{#1}}}}
\newcommand{\ipder}[3][]{\partial^{#1}{#2}/\partial{#3^{#1}}}
\newcommand{\dder}[3]{\frac{\partial^2{#1}}{\partial{#2}\partial{#3}}}

\newcommand{\wb}[1]{\overline{#1}}
\newcommand{\wt}[1]{\widetilde{#1}}

\def\xssp{\hspace{1pt}}
\def\ssp{\hspace{3pt}}
\def\msp{\hspace{5pt}}
\def\lsp{\hspace{12pt}}

\newcommand{\cA}{\mathcal{A}}
\newcommand{\cB}{\mathcal{B}}
\newcommand{\cC}{\mathcal{C}}
\newcommand{\cD}{\mathcal{D}}
\newcommand{\cE}{\mathcal{E}}
\newcommand{\cF}{\mathcal{F}}
\newcommand{\cG}{\mathcal{G}}
\newcommand{\cH}{\mathcal{H}}
\newcommand{\cI}{\mathcal{I}}
\newcommand{\cJ}{\mathcal{J}}
\newcommand{\cK}{\mathcal{K}}
\newcommand{\cL}{\mathcal{L}}
\newcommand{\cM}{\mathcal{M}}
\newcommand{\cN}{\mathcal{N}}
\newcommand{\cO}{\mathcal{O}}
\newcommand{\cP}{\mathcal{P}}
\newcommand{\cQ}{\mathcal{Q}}
\newcommand{\cR}{\mathcal{R}}
\newcommand{\cS}{\mathcal{S}}
\newcommand{\cT}{\mathcal{T}}
\newcommand{\cU}{\mathcal{U}}
\newcommand{\cV}{\mathcal{V}}
\newcommand{\cW}{\mathcal{W}}
\newcommand{\cX}{\mathcal{X}}
\newcommand{\cY}{\mathcal{Y}}
\newcommand{\cZ}{\mathcal{Z}}

\newcommand{\vA}{\mathbf{A}}
\newcommand{\vB}{\mathbf{B}}
\newcommand{\vC}{\mathbf{C}}
\newcommand{\vD}{\mathbf{D}}
\newcommand{\vE}{\mathbf{E}}
\newcommand{\vF}{\mathbf{F}}
\newcommand{\vG}{\mathbf{G}}
\newcommand{\vH}{\mathbf{H}}
\newcommand{\vI}{\mathbf{I}}
\newcommand{\vJ}{\mathbf{J}}
\newcommand{\vK}{\mathbf{K}}
\newcommand{\vL}{\mathbf{L}}
\newcommand{\vM}{\mathbf{M}}
\newcommand{\vN}{\mathbf{N}}
\newcommand{\vO}{\mathbf{O}}
\newcommand{\vP}{\mathbf{P}}
\newcommand{\vQ}{\mathbf{Q}}
\newcommand{\vR}{\mathbf{R}}
\newcommand{\vS}{\mathbf{S}}
\newcommand{\vT}{\mathbf{T}}
\newcommand{\vU}{\mathbf{U}}
\newcommand{\vV}{\mathbf{V}}
\newcommand{\vW}{\mathbf{W}}
\newcommand{\vX}{\mathbf{X}}
\newcommand{\vY}{\mathbf{Y}}
\newcommand{\vZ}{\mathbf{Z}}

\newcommand{\va}{\mathbf{a}}
\newcommand{\vb}{\mathbf{b}}
\newcommand{\vc}{\mathbf{c}}
\newcommand{\vd}{\mathbf{d}}
\newcommand{\ve}{\mathbf{e}}
\newcommand{\vf}{\mathbf{f}}
\newcommand{\vg}{\mathbf{g}}
\newcommand{\vh}{\mathbf{h}}
\newcommand{\vi}{\mathbf{i}}
\newcommand{\vj}{\mathbf{j}}
\newcommand{\vk}{\mathbf{k}}
\newcommand{\vl}{\mathbf{l}}
\newcommand{\vm}{\mathbf{m}}
\newcommand{\vn}{\mathbf{n}}
\newcommand{\vo}{\mathbf{o}}
\newcommand{\vp}{\mathbf{p}}
\newcommand{\vq}{\mathbf{q}}
\newcommand{\vr}{\mathbf{r}}
\newcommand{\Vs}{\mathbf{s}}
\newcommand{\vt}{\mathbf{t}}
\newcommand{\vu}{\mathbf{u}}
\newcommand{\vv}{\mathbf{v}}
\newcommand{\vw}{\mathbf{w}}
\newcommand{\vx}{\mathbf{x}}
\newcommand{\vy}{\mathbf{y}}
\newcommand{\vz}{\mathbf{z}}

\newcommand{\vone}{\mathbf{1}}
\newcommand{\vzero}{\mathbf{0}}

\newcommand{\valpha}{{\boldsymbol{\alpha}}}
\newcommand{\vbeta}{{\boldsymbol{\beta}}}
\newcommand{\vgamma}{{\boldsymbol{\gamma}}}
\newcommand{\vdelta}{{\boldsymbol{\delta}}}
\newcommand{\vepsilon}{{\boldsymbol{\epsilon}}}
\newcommand{\vzeta}{{\boldsymbol{\zeta}}}
\newcommand{\veta}{{\boldsymbol{\eta}}}
\newcommand{\vtheta}{{\boldsymbol{\theta}}}
\newcommand{\viota}{{\boldsymbol{\iota}}}
\newcommand{\vkappa}{{\boldsymbol{\kappa}}}
\newcommand{\vlambda}{{\boldsymbol{\lambda}}}
\newcommand{\vmu}{{\boldsymbol{\mu}}}
\newcommand{\vnu}{{\boldsymbol{\nu}}}
\newcommand{\vxi}{{\boldsymbol{\xi}}}
\newcommand{\vomikron}{{\boldsymbol{\omikron}}}
\newcommand{\vpi}{{\boldsymbol{\pi}}}
\newcommand{\vrho}{{\boldsymbol{\rho}}}
\newcommand{\vsigma}{{\boldsymbol{\sigma}}}
\newcommand{\vtau}{{\boldsymbol{\tau}}}
\newcommand{\vupsilon}{{\boldsymbol{\upsilon}}}
\newcommand{\vphi}{{\boldsymbol{\phi}}}
\newcommand{\vchi}{{\boldsymbol{\chi}}}
\newcommand{\vpsi}{{\boldsymbol{\psi}}}
\newcommand{\vomega}{{\boldsymbol{\omega}}}

\newcommand{\rLambda}{\mathrm{\Lambda}}
\newcommand{\rSigma}{\mathrm{\Sigma}}

\newcommand{\vLambda}{\bm{\rLambda}}
\newcommand{\vSigma}{\bm{\rSigma}}

\makeatletter
\newcommand*\bdot{\mathpalette\bdot@{.7}}
\newcommand*\bdot@[2]{\mathbin{\vcenter{\hbox{\scalebox{#2}{$\m@th#1\bullet$}}}}}
\makeatother

\makeatletter
\DeclareRobustCommand\onedot{\futurelet\@let@token\@onedot}
\def\@onedot{\ifx\@let@token.\else.\null\fi\xspace}

\def\eg{\emph{e.g}\onedot} \def\Eg{\emph{E.g}\onedot}
\def\ie{\emph{i.e}\onedot} \def\Ie{\emph{I.e}\onedot}
\def\cf{\emph{cf}\onedot} \def\Cf{\emph{Cf}\onedot}
\def\etc{\emph{etc}\onedot} \def\vs{\emph{vs}\onedot}
\def\wrt{w.r.t\onedot} \def\dof{d.o.f\onedot} \def\aka{a.k.a\onedot}
\def\etal{\emph{et al}\onedot}
\makeatother

\newcommand{\base}{\mathrm{base}}
\newcommand{\novel}{\mathrm{novel}}
\newcommand{\NN}{\mathrm{NN}}
\newcommand{\masked}{\mathrm{masked}}
\newcommand{\soft}{\mathrm{soft}}

\newcommand{\ours}{iLPC\xspace}
\newcommand{\lp}{LP\xspace}
\newcommand{\bal}{Balance\xspace}
\newcommand{\ilc}{iLC\xspace}
\newcommand{\cp}{Class\xspace}
\newcommand{\iprob}{iProb\xspace}
\newcommand{\pt}{PT\xspace}

\maketitle
\ificcvfinal\thispagestyle{empty}\fi

\begin{abstract}
Few-shot learning amounts to learning representations and acquiring knowledge such that novel tasks may be solved with both supervision and data being limited. Improved performance is possible by transductive inference, where the entire test set is available concurrently, and semi-supervised learning, where more unlabeled data is available.

Focusing on these two settings, we introduce a new algorithm that leverages the manifold structure of the labeled and unlabeled data distribution to predict pseudo-labels, while balancing over classes and using the loss value distribution of a limited-capacity classifier to select the cleanest labels, iteratively improving the quality of pseudo-labels. Our solution surpasses or matches the state of the art results on four benchmark datasets, namely \emph{mini}ImageNet, \emph{tiered}ImageNet, CUB and CIFAR-FS, while being robust over feature space pre-processing and the quantity of available data. The publicly available source code can be found in \url{https://github.com/MichalisLazarou/iLPC}.
\end{abstract}


\section{Introduction}
\label{sec:intro}

\emph{Few-shot learning}~\cite{matchingNets,prototypical} is challenging the deep learning paradigm in that, not only supervision is limited, but data is limited too. Despite the initial promise of \emph{meta-learning}~\cite{SNAIL,MAML}, \emph{transfer learning}~\cite{closerlook,rfs} is becoming increasingly successful in decoupling representation learning from learning novel tasks on limited data. \emph{Semi-supervised learning}~\cite{Lee13,mixmatch} is one of the dominant ways of dealing with limited supervision and indeed, its few-shot learning counterparts~\cite{ssmeta,transmatch} are miniature versions where both labeled and unlabeled data are limited proportionally, while representation learning may be decoupled. These methods are closer to \emph{transductive inference}~\cite{tpn,embeddingpropagation}, which
was a pillar of semi-supervised learning before deep learning~\cite{semibook}.

\begin{figure}
\centering
\extfig{idea}{
\begin{tikzpicture}[
	scale=.13,
	font=\scriptsize,
	r/.style = {fill=red},
	g/.style = {fill=green!70!black},
	b/.style = {fill=orange!80},
	net/.style = {draw=green!60!black,fill=green!30},
	trap/.style = {inner sep=6pt,trapezium,trapezium angle=70,shape border uses incircle,shape border rotate=-90},
	ex/.style = {draw,circle,inner sep=1.4pt,outer sep=0},
	nex/.style = {every node/.style=ex},
	num/.style = {draw,circle,inner sep=1pt,outer sep=0pt},
	n/.style = {draw=none},
	o9/.style = {fill opacity=.9},
	o8/.style = {fill opacity=.8},
	o7/.style = {fill opacity=.7},
	o6/.style = {fill opacity=.6},
	o5/.style = {fill opacity=.5},
	o4/.style = {fill opacity=.4},
	o3/.style = {fill opacity=.3},
	o2/.style = {fill opacity=.2},
	o1/.style = {fill opacity=.1},
]
	\matrix[
		row sep=3pt,column sep=12pt,cells={scale=.13},
	]
	{
	\&
		\node at(2,0) (1){feature \\ extraction};
		\node[num,left=1pt of 1]{1};
	\&
		\node at(2,0) (2){nearest \\ neighbor \\ graph};
		\node[num,left=1pt of 2]{2};
	\&
		\node at(2,0) (3){label \\ propagation};
		\node[num,left=1pt of 3]{3};
	\\[-18pt]
		\draw    (0,-4)  rectangle ++(10,1);
		\draw    (0,-6)  rectangle ++(10,1) +(1,-.5) coordinate(q1);
		\draw    (0,-8)  rectangle ++(10,1);
		\draw[r] (0, 2)  rectangle ++(10,1);
		\draw[g] (0, 4)  rectangle ++(10,1) +(1,-.5) coordinate(s1);
		\draw[b] (0, 6)  rectangle ++(10,1);
		\node at(5,-1.5) {queries $Q$};
		\node at(5, 8.5) {support $S$};
	\&
		\node[net,trap] (net){$f$};
		\path (net.west) +(-1,4) coordinate(s2) +(-1,-4) coordinate(q2);
		\path (net.east) +(1,0) coordinate(g1);
	\&
		\path[nex] 
			( 0, 0) node   (x1) {} ( 3, 1) node(x2) {} ( 1, 3) node(x3) {} (2,5) node(x4){}
			( 4, 3) node[g](x5) {} ( 6, 1) node(x6) {} ( 7, 4) node(x7) {}
			(-2, 4) node[b](x8) {} (-5, 3) node(x9) {} (-4, 6) node(x10){}
			(-1,-3) node[r](x11){} (-3,-2) node(x12){} (-1,-6) node(x13){}
			( 4,-2) node   (x14){} ( 2,-4) node(x15){} ( 6,-5) node(x16){};
		\draw
			(x1)--(x2)--(x5)--(x6)--(x7)--(x5)--(x4)--(x3)--(x1)
			(x3)--(x8)--(x9)--(x12)--(x1)
			(x12)--(x11)--(x15)--(x14)--(x2)
			(x8)--(x10) (x11)--(x13) (x15)--(x16);
		\coordinate (g2) at(-5,0);
		\coordinate (l1) at(7,0);
		\coordinate (i2) at(-5,-4);
	\&
		\path[nex] 
			( 0, 0) node      (y1) {} ( 3, 1) node[g,o2](y2) {} ( 1, 3) node[b,o2](y3) {} (2,5) node[g,o2](y4){}
			( 4, 3) node[g]   (y5) {} ( 6, 1) node[g,o5](y6) {} ( 7, 4) node[g,o5](y7) {}
			(-2, 4) node[b]   (y8) {} (-5, 3) node[b,o5](y9) {} (-4, 6) node[b,o5](y10){}
			(-1,-3) node[r]   (y11){} (-3,-2) node[r,o5](y12){} (-1,-6) node[r,o5](y13){}
			( 4,-2) node[r,o2](y14){} ( 2,-4) node[r,o5](y15){} ( 6,-5) node(y16){};
		\draw
			(y1)--(y2)--(y5)--(y6)--(y7)--(y5)--(y4)--(y3)--(y1)
			(y3)--(y8)--(y9)--(y12)--(y1)
			(y12)--(y11)--(y15)--(y14)--(y2)
			(y8)--(y10) (y11)--(y13) (y15)--(y16);
		\coordinate (l2) at(-5,0);
		\coordinate (b1) at(0,-7);
	\\[12pt]
	\&
		\path[nex]
			(-4, 4) node[b]{} ++(4,0) node[r]{} ++(4,0) node[g]{}
			(-4, 2) node[b]{} ++(4,0) node[r]{} ++(4,0) node[g]{}
			(-4,-1) node   {} ++(4,0) node   {} ++(4,0) node   {}
			(-4,-3) node   {} ++(4,0) node   {} ++(4,0) node   {};
		\draw 
			(-5.5,5.5) rectangle +(11,-5)
			++(0,-5)   rectangle +(11,-5);
		\coordinate (a2) at(6.5,0);
		\coordinate (i1) at(4,6.5);
	\&
		\path[nex]
			(-4, 4) node[  b   ]{} ++(4,0) node[  r   ]{} ++(4,0) node[  g   ]{}
			(-4, 1) node[  b   ]{} ++(4,0) node[  r   ]{} ++(4,0) node[  g   ]{}
			(-4,-1) node[n,b,o7]{} ++(4,0) node[n,r,o7]{} ++(4,0) node[n,g,o7]{}
			(-4,-3) node[n,b,o4]{} ++(4,0) node[n,r,o4]{} ++(4,0) node[n,g,o4]{};
		\draw 
			(-5.5,5.5) rectangle +(11,-3)
			++(0,-3)   rectangle +(11,-7);
		\coordinate (c2) at(6.5,0);
		\coordinate (a1) at(-6.5,0);
	\&
		\path[nex]
			(-4, 4) node[  b   ]{} ++(4,0) node[  r   ]{} ++(4,0) node[  g   ]{}
			(-4, 1) node[n,b,o7]{} ++(4,0) node[n,r   ]{} ++(4,0) node[n,g,o4]{}
			(-4,-1) node[n,b   ]{} ++(4,0) node[n,r,o4]{} ++(4,0) node[n,g,o7]{}
			(-4,-3) node[n,b,o4]{} ++(4,0) node[n,r,o7]{} ++(4,0) node[n,g   ]{};
		\draw 
			(-5.5,5.5) rectangle +(11,-3)
			++(0,-3)   rectangle +(11,-7);
		\coordinate (b2) at(0,6.5);
		\coordinate (c1) at(-6.5,0);
	\\
	\&
		\node at(2,0) (6){support set \\ augmentation};
		\node[num,left=1pt of 6]{6};
	\&
		\node at(2,0) (5){label \\ cleaning};
		\node[num,left=1pt of 5]{5};
	\&
		\node at(2,0) (4){class \\ balancing};
		\node[num,left=1pt of 4]{4};
	\\
	};
	\draw[->] (s1)--(s2);
	\draw[->] (q1)--(q2);
	\draw[->] (g1)--(g2);
	\draw[->] (l1)--(l2);
	\draw[->] (b1)--(b2);
	\draw[->] (c1)--(c2);
	\draw[->] (a1)--(a2);
	\draw (i1) edge[->,out=60,in=200] coordinate[midway](it) (i2);
	\node[num,below right=4pt and 2pt of it] (7){7};
	\node[right=1pt of 7] {iteration};
\end{tikzpicture}
}
\caption{Overview of the proposed method. See text for details.}
\label{fig:idea}
\end{figure}
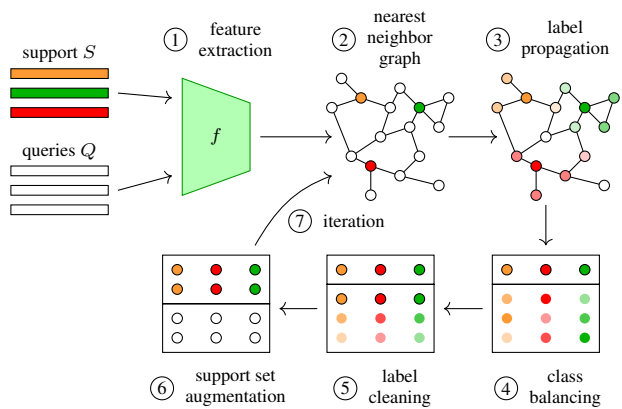

Predicting \emph{pseudo-labels} on unlabeled data~\cite{Lee13} is one of the oldest ideas in semi-supervised learning~\cite{scudder1965probability}. Graph-based methods, in particular \emph{label propagation}~\cite{ZhGh02,ZBL+03}, are prominent in transductive inference and translate to inductive inference in deep learning exactly by predicting pseudo-labels~\cite{semilpavrithis}. However, with the representation being fixed, the quality of pseudo-labels is critical in few-shot learning~\cite{lrici,mct}. At the same time, in learning with noisy labels~\cite{ArazoOAOM19,o2u,robustonfly}, it is common to clean labels based on the loss value statistics of a small-capacity classifier.

In this work, we leverage these ideas to improve transductive and semi-supervised few-shot learning. As shown in \autoref{fig:idea}, focusing on transduction, a set of labeled support examples $S$ and unlabeled queries $Q$ are given, represented in a feature space by mapping $f$. By label propagation~\cite{ZBL+03}, we obtain a matrix that associates examples to classes. The submatrix corresponding to unlabeled examples, $P$, is normalized over examples and classes using the Sinkhorn-Knopp algorithm~\cite{sinkhornknopp}, assuming a uniform distribution over classes. We extract pseudo-labels from $P$, which we clean following O2U-Net~\cite{o2u}, keeping only one example per class. Finally, inspired by~\cite{KoDi12}, we move these examples from $Q$ to $S$ and iterate until $Q$ is empty.

\section{Related work and contributions}
\label{sec:related}



\subsection{Few-shot learning}

\paragraph{Meta-learning}

This is a popular paradigm, where the training set is partitioned in episodes resembling in structure the novel tasks~\cite{MAML, siamese, prototypical, SNAIL, matchingNets}. \emph{Model-based} methods rely on the properties of specific model architectures, such as recurrent and memory-augmented networks~\cite{SNAIL, santoro, metaNet}. \emph{Optimization-based} methods attempt to learn model parameters that are able to adapt fast in novel tasks~\cite{MAML, iMAML, CAVIA, ravilstm, reptile, metaoptnet, bertinetto}. \emph{Metric-based} methods attempt to learn representations that are appropriate for comparisons~\cite{prototypical, siamese, relationnet, matchingNets}. Of course, \emph{metric learning} is a research area on its own~\cite{WHH+19,KKCK20} and modern ideas are commonly effective in few-shot learning~\cite{ECCV2020_162}.

\paragraph{Predicting weights, data augmentation}

Also based on meta-learning, it is possible to predict new parameters or even data. For instance, it is common to learn to predict data-dependent network parameters in the last layer (classifier)~\cite{gidarisFSL,imprintedweights,activationweights} or even in intermediate convolutional layers~\cite{bertinetto2016learning}. Alternatively, one can learn to generate novel-task data in the feature space~\cite{lowshot2} or in the input (image) space~\cite{lowshot,dataAG}. The quality can be improved by translating images, similar to style transfer~\cite{liu2019few}. Such learned data augmentation is complementary to other ideas.

\paragraph{Transfer learning}

More recently, it is recognized that learning a powerful representation on the entire training set is more effective than sampling few-shot training episodes that resemble novel tasks~\cite{gidarisFSL,closerlook,rfs,manifoldmixup}. In doing so, one may use standard loss functions~\cite{gidarisFSL,closerlook}, knowledge distillation~\cite{rfs} or other common self-supervision and regularization methods~\cite{manifoldmixup}. We follow this \emph{transfer learning} approach, which allows us to decouple representation learning from the core few-shot learning idea and provide clearer comparisons with the competition.


\subsection{Using unlabeled data} 

Leveraging unlabelled data is of interest due to the ease of obtaining such data. Two common settings are transductive inference and semi-supervised learning.

\paragraph{Transductive inference} 

In this setting, all novel-class unlabeled query examples are assumed available at the same time at inference~\cite{tpn,tafssl,transductiveEA,exploitingUI,pt_map,embeddingpropagation}. These examples give additional information on the distribution of novel classes on top of labeled support examples.

Common transductive inference solutions are adapted for few-shot classification, notably \emph{label propagation}~\cite{tpn} and \emph{embedding propagation}~\cite{embeddingpropagation}, which smooths embeddings as in image segmentation~\cite{Bertasius_2017_CVPR}. Both operations are also used at representation learning, as in meta-learning. Using dimensionality reduction, TAFSSL~\cite{tafssl} learns a task-specific feature subspace that is highly discriminant for novel tasks. \emph{Meta-confidence transduction} (MCT)~\cite{mct} meta-learns a data-dependent scaling function to normalize every example and iteratively updates class centers. PT+MAP~\cite{pt_map} uses a similar iterative process but also balances over classes. \emph{Cross-attention}~\cite{crossattention}, apart from aligning feature maps by correlation, leverages query examples by iteratively making predictions and using the most confident ones to update the class representation.

\paragraph{Semi-supervised learning} 

In this case, labeled novel-class support examples and additional unlabeled data
are given. A classifier may be learned on both to make predictions on novel-class queries~\cite{ssmeta,lrici,transmatch}.

One of the first contributions uses unlabelled examples to adapt prototypical networks~\cite{prototypical}, while discriminating from distractor classes~\cite{ssmeta}. Common semi-supervised solutions are also adapted to few-shot classification, for instance \emph{learning to self-train}~\cite{learning2selftrain}, which adapts \emph{pseudo-label}~\cite{Lee13} and TransMatch~\cite{transmatch}, which is an adaptation of MixMatch~\cite{mixmatch}. \emph{Instance credibility inference}~\cite{lrici} predicts pseudo-labels iteratively, using a linear classifier to select the most likely to be correct and then augmenting the support set. \emph{Adaptive subspaces}~\cite{adaptivesubspace} are learned from labeled and unlabeled data, yielding a discriminative subspace classifier that maximizes the margin between subspaces.

\subsection{Contributions} 

In this work, focusing on the transfer learning paradigm to learn novel tasks given a fixed representation~\cite{rfs,manifoldmixup}, we make the following contributions:

\begin{enumerate}[itemsep=2pt, parsep=0pt, topsep=0pt]
	\item We combine the power of predicting pseudo-labels in semi-supervised learning~\cite{Lee13,learning2selftrain} with label cleaning in learning from noisy labels~\cite{o2u}.
	\item According to manifold assumption, we use label propagation~\cite{ZBL+03,tpn} to infer pseudo-labels, while balancing over classes~\cite{sinkhornknopp,pt_map} and iteratively re-use pseudo-labels in the propagation process~\cite{KoDi12,lrici}.
	\item We achieve new state of the art in both transductive and semi-supervised few-shot learning.
\end{enumerate}

\section{Method}
\label{sec:method}



\subsection{Problem formulation}
\label{sec:problem}


At \emph{representation learning}, we assume access to a labeled dataset $D_{\base}$ with each example having a label in one of the classes in $C_{\base}$. This dataset is used to learn a mapping $f: \cX \to \real^d$ from an input space $\cX$ to a $d$-dimensional \emph{feature} or \emph{embedding} space.

The knowledge acquired at representation learning is used to solve \emph{novel tasks}, assuming access to a dataset $D_{\novel}$ with each example being associated with one of the classes $C_{\novel}$, where $C_{\novel}$ is disjoint from $C_{\base}$. Examples in $D_{\novel}$ may be labeled or not.

In \emph{few-shot classification}~\cite{matchingNets}, a novel task is defined by sampling a \emph{support set} $S$ from $D_{\novel}$, consisting of $N$ classes with $K$ labeled examples per class, for a total of $L \defn NK$ examples. Given the mapping $f$ and the support set $S$, the problem is to learn an $N$-way classifier that makes predictions on unlabeled queries also sampled from $D_{\novel}$. 
Queries are treated independently of each other. This is referred to as \emph{inductive inference}.

In \emph{transductive inference}, a \emph{query set} $Q$ consisting of $M$ unlabeled examples is also sampled from $D_{\novel}$. Given the mapping $f$, $S$ and $Q$, the problem is to make predictions on $Q$, without necessarily learning a classifier. In doing so, one may leverage the distribution of examples in $Q$, which is important because $M$ is assumed greater than $L$.

In \emph{semi-supervised} few-shot classification, an unlabelled set $U$ of $M$ unlabeled examples is also sampled from $D_{\novel}$. Given $f$, $S$ and $U$, the problem is to learn to make predictions on new queries from $D_{\novel}$, as in inductive inference. Again, $M > L$ and we may leverage the distribution of $U$.

In this work, we focus on transductive inference and semi-supervised classification, given $f$. The performance of $f$ on inductive inference is one of our baselines. We develop our solution for transductive inference. In the semi-supervised case, we follow the same solution with $Q$ replaced by $U$. Using the predictions on $U$, we then proceed as in the inductive case, with $S$ replaced by $S \cup U$.


\subsection{Nearest-neighbor graph}
\label{seq:graph}

We are given the mapping $f$, the labeled support set $S \defn \{ (x_i,y_i) \}_{i=1}^L$ and the query set $Q \defn \{ x_{L+i} \}_{i=1}^M$, where $y_i \in [N] \defn \{1,\dots,N\}$. We embed all examples from $S$ and $Q$ into $V = \{\vv_1, \dots, \vv_T\} \subset \real^d$ and $\ell_1$-normalize them, where $T \defn L + M$ and $\vv_i \defn f(x_i)$ for $i \in [T]$. Following~\cite{semilpavrithis}, we construct a $k$-nearest neighbour graph of the features in $V$, represented by a sparse $T \times T$ nonnegative \emph{affinity matrix} $A$, with
\begin{equation}
	A_{ij} \defn
	\begin{cases}
		[\vv_i\tran \vv_j]_+^\gamma, & \mif i \ne j \wedge \vv_i \in \NN_k(\vv_j) \\
		0,                           & \other
	\end{cases}
\label{eq:affinity}
\end{equation}
for $i \in [T]$, $j \in [N]$, where $\NN_k(\vv)$ are the $k$-nearest neighbors of $\vv$ in $V$ and $\gamma > 1$ is a hyperparameter. Finally, we obtain the symmetric $T \times T$ \emph{adjacency matrix} $W \defn \frac{1}{2} (A + A\tran)$ and we symmetrically normalize it as 
\begin{equation}
    \cW \defn D^{-1/2} W D^{-1/2},
\label{eq:adj}
\end{equation}
where $D = \diag(W \vone_T)$ is the $T \times T$ degree matrix of $W$.


\subsection{Label propagation}
\label{sec:lp}

Following~\cite{ZBL+03}, we define the $T \times N$ \emph{label matrix} $Y$ as
\begin{equation}
 Y_{ij} \defn
	\begin{cases}
		1, & \mif i \le L \wedge y_i = j \\
		0, & \other
	 \end{cases}
\label{eq:label}
\end{equation}
for $i \in [T]$, $j \in [N]$. Matrix $Y$ has one column per class and one row per example, which is an one-hot encoded label for 
$S$ and a zero vector for 
$Q$. Label propagation amounts to solving $N$ linear systems
\begin{equation}
    Z \defn (I-\alpha \cW)^{-1} Y,
\label{eq:lp}
\end{equation}
where $\alpha \in [0, 1)$ is a hyperparameter. The resulting $T \times N$ matrix $Z$ 
can be used to make predictions by taking the maximum element per row~\cite{ZBL+03}. However, before making predictions, we balance over classes.


\subsection{Class balancing}
\label{sec:balance}

We focus on the $M \times N$ submatrix 
\begin{equation}
    P \defn Z_{L+1:T,:}
\label{eq:sub}
\end{equation}
(the last $M$ rows) of $Z$ that corresponds to unlabeled queries. We
first perform an element-wise \emph{power transform}
\begin{equation}
    P_{ij} \gets P_{ij}^\tau
\label{eq:power}
\end{equation}
for $i \in [M]$, $j \in [N]$, where $\tau > 1$, encouraging hard predictions.
Parameter $\tau$ is analogous to the \emph{scale} (or \emph{inverse temperature}) of logits in softmax-based classifiers~\cite{gidarisFSL,imprintedweights,tadam}, only here the elements of $P$ are proportional to class probabilities rather than logits.

Inspired by~\cite{pt_map}, we normalize $P$ to a given row-wise sum $\vp \in \real^M$ and column-wise sum $\vq \in \real^N$. Each element $p_i \in [0,1]$ of $\vp$ represents a confidence of example $x_{L+i}$ for $i \in [M]$; it can be a function of the $i$-th row of $P$ or set to $1$. Each element $q_j \ge 0$ of $\vq$ represents a weight of class $j$ for $j \in [N]$. In the absence of such information, we set 
\begin{equation}
    \vq \defn \frac{1}{N} (\vp\tran \vone_M) \vone_N,
\label{eq:balance}
\end{equation}
assuming a uniform distribution of queries over classes.

The normalization itself is a projection of $P$ onto the set $\mathbb{S}(\vp,\vq)$ of nonnegative $M \times N$ matrices having row-wise sum $\vp$ and column-wise sum $\vq$,
\begin{equation}
    \mathbb{S}(\vp,\vq) \defn \{ X \in \real^{M \times N}: X \vone_N = \vp, X\tran \vone_M = \vq \}.
\label{eq:norm}
\end{equation}
We use the \emph{Sinkhorn-Knopp} algorithm~\cite{sinkhornknopp} for this projection, which alternates between rescaling the rows of $P$ to sum to $\vp$ and its columns to sum to $\vq$,
\begin{align}
    P & \gets \diag(\vp) \diag(P \vone_N)^{-1} P       \label{eq:sink-row} \\
    P & \gets P \diag(P\tran \vone_M)^{-1} \diag(\vq), \label{eq:sink-col}
\end{align}
until convergence. Finally, for each query $x_{L+i}$, $i \in [M]$, we predict the \emph{pseudo-label}
\begin{equation}
	 \hat{y}_{L+i} \defn \arg\max_{j \in [N]} P_{ij}
\label{eq:argmax}
\end{equation}
that corresponds to the maximum element of the $i$-th row of the resulting matrix $P$, for $i \in [M]$.


\subsection{Label cleaning}
\label{sec:clean}

The predicted pseudo-labels are not necessarily correct, yet a classifier can be robust to such noise. This is the case when enough data is available to adapt the representation~\cite{Lee13,semilpavrithis}, such that the quality of pseudo-labels improves with training. Since data is limited here, we would like to select pseudo-labeled queries in $Q$ that are most likely to be correct, treat them as truly labeled and add them to the support set $S$. Iterating this process is an alternative way of improving the quality of pseudo-labels.

We interpret this problem as \emph{learning with noisy labels}, leveraging recent advances in label cleaning~\cite{ArazoOAOM19,o2u,robustonfly}. Assuming that the classifier does not overfit the data, \eg with small capacity, high learning rate or few iterations, the principle is that examples with clean labels exhibit less loss than examples with noisy labels.

In particular, given the labeled support set $S \defn \{ (x_i,y_i) \}_{i=1}^L$ and the pseudo-labeled query set $\hat{Q} \defn \{ (x_{L+i},\hat{y}_{L+i}) \}_{i=1}^M$, we train an $N$-way classifier $g$ using a weighted cross-entropy loss
\begin{equation}
	\ell \defn - \sum_{i=1}^L \log g(x_i)_{y_i} - \sum_{i=1}^M p_i \log g(x_{L+i})_{\hat{y}_{L+i}},
\label{eq:loss}
\end{equation}
where $p_i$ is the confidence weight of example $x_{L+i}$. Here, the classifier $g$ is assumed to yield a vector of probabilities over classes using softmax and $g(x)_y$ refers to element $y \in [N]$ of $g(x)$. In practice, it is obtained by a linear classifier on top of embedding $f$, optionally allowing the adaptation of the last layers of the network implementing $f$.

The loss term $\ell_i \defn -p_i \log g(x_{L+i})_{\hat{y}_{L+i}}$, corresponding to the pseudo-labeled query $x_{L+i}$, is used for selection. Following O2U-Net~\cite{o2u}, we use large learning rate and collect the average loss $\bar{l}_i$ over all epochs, for $i \in [M]$. In learning with noisy labels, it is common to detect noisy labels based on statistics of this loss for clean and noisy labels~\cite{ArazoOAOM19,robustonfly}. However, this does not work well with predicted pseudo-labels~\cite{albert2020relab}, hence we select queries having the least average loss~\cite{o2u,albert2020relab}. 
The extreme case of selecting one query example per class is defined as
\begin{equation}
	\cI \defn \left\{ \arg\min_{\hat{y}_{L+i} = j} \bar{\ell}_i: j \in [N] \right\}.
\label{eq:least}
\end{equation}
Finally, we augment the support set $S$ with the selected queries and their pseudo-labels, while at the same time removing the selected queries from $Q$.
\begin{align}
	S & \gets S \cup \{ (x_{L+i},\hat{y}_{L+i}) \}_{i \in \cI} \label{eq:aug-support} \\
	Q & \gets Q \setminus \{ x_{L+i} \}_{i \in \cI}            \label{eq:aug-query}
\end{align}


\subsection{Iterative inference}
\label{sec:iter}

Although label propagation and class balancing
make predictions on the entire unlabeled query set $Q$, we apply cleaning to keep $\nu$ pseudo-labeled query per class, which we move from $Q$ to the support set $S$.
We iterate the entire process, selecting $\nu$ pseudo-labeled queries per class at a time, until $Q$ is empty and $S$ is augmented with all pseudo-labeled queries. Assuming that the selections
are 
correct, the idea is that treating them as truly labeled in $S$ improves the quality of the pseudo-labels.

\begin{algorithm}
\footnotesize

\DontPrintSemicolon
\SetFuncSty{textsc}
\SetDataSty{emph}
\newcommand{\commentsty}[1]{{\color{DarkGreen}#1}}
\SetCommentSty{commentsty}
\SetKwComment{Comment}{$\triangleright$ }{}

\SetKwInOut{Input}{input}
\SetKwInOut{Output}{output}
\SetKwFunction{Graph}{graph}
\SetKwFunction{Label}{label}
\SetKwFunction{LP}{lp}
\SetKwFunction{Power}{power}
\SetKwFunction{Balance}{balance}
\SetKwFunction{Sinkhorn}{Sinkhorn}
\SetKwFunction{Predict}{predict}
\SetKwFunction{Clean}{clean}
\SetKwFunction{Augment}{augment}

\Input{ embedding $f$}
\Input{ labeled support set $S$ with $\card{S}=L$}
\Input{ unlabeled query set $Q$ with $\card{Q}=M$}
\Output{ augmented support set $S$ with $\card{S}=L+M$}
\BlankLine

\Repeat{$Q = \emptyset$ \Comment*[f]{all queries are predicted}}
{
	$\cW \gets \Graph(f,S,Q;\gamma,k)$ \Comment*{adjacency matrix~\eq{affinity},\eq{adj}}
	$Y \gets \Label(S)$ \Comment*{label matrix~\eq{label}}
	$Z \gets \LP(\cW,Y;\alpha)$ \Comment*{label propagation~\eq{lp}}
	$P \gets Z_{L+1:L+M,:}$ \Comment*{unlabeled submatrix~\eq{sub}}
	$P \gets \Power(P;\tau)$ \Comment*{power transform~\eq{power}}
	$(\vp,\vq) \gets \Balance(P)$ \Comment*{class balance~\eq{balance}}
	$P \gets \Sinkhorn(P;\vp,\vq)$ \Comment*{Sinkhorn-Knopp~\eq{sink-row},\eq{sink-col}}
	$\hat{Y} \gets \Predict(P)$ \Comment*{pseudo-labels~\eq{argmax}}
	$\cI \gets \Clean(f,S,Q,\hat{Y},\vp)$ \Comment*{label cleaning~\eq{loss},\eq{least}}
	$(S,Q) \gets \Augment(S,Q,\cI)$ \Comment*{augment support~\eq{aug-support},\eq{aug-query}}
}

\caption{Iterative label propagation and cleaning (\ours).}
\label{alg:main}
\end{algorithm}

Algorithm~\ref{alg:main} summarizes this process, called \emph{iterative label propagation and cleaning} (\ours). Given $S$, $Q$ and the embedding $f$, we construct the nearest neighbor graph represented by the normalized adjacency matrix $\cW$~\eq{affinity},\eq{adj} and we perform label propagation on the current label matrix $Y$~\eq{lp}. Focusing on the unlabeled submatrix $P$ of the resulting matrix $Z$, we perform power transform~\eq{power} and row/column normalization to balance over classes~\eq{sink-row},\eq{sink-col}. We predict pseudo-labels $\hat{Y}$ from the normalized $P$~\eq{argmax}, which we use along with $S$ and $Q$ to train a linear classifier on top of $f$ with cross entropy loss~\eq{loss} and a cyclical learning rate schedule~\cite{o2u}. We select one query per class with the least average loss over all epochs~\eq{least},
 which we move from $Q$ to $S$ as labeled~\eq{aug-support},\eq{aug-query}.
With $Q,S$ redefined, we repeat the process until $Q$ is empty. 

At termination, all data is labeled in $S$. The predicted labels over the original queries are the output in the case of transductive inference. In semi-supervised classification, we use $S$ to learn a new classifier and make predictions on new queries, as in inductive inference.


\section{Experiments}
\label{sec:experiments}

\input{fig/data/sample}
\newcommand{\ci}[1]{{\tiny $\pm$#1}}
\newcommand{\cip}{\phantom{\ci{0.00}}}
\newcommand{\cim}{\ci{\alert{0.00}}}



\begin{table*}
\small
\centering
\begin{tabular}{lcccccccccc}
\toprule
\Th{Inference}          & \mc{5}{\Th{Components}}          & \mc{2}{\Th{ResNet-12A}}                   & \mc{2}{\Th{WRN-28-10}}                    \\ 
                        & \lp & \bal & \ilc & \iprob & \cp & 1-shot              & 5-shot              & 1-shot              & 5-shot              \\ \midrule
Inductive               &     &      &      &        &     & 56.30\ci{0.62}      & 75.59\ci{0.47}      & 68.17\ci{0.60}      & 84.33\ci{0.43}      \\ 
Transductive            & \ch &      &      &        &     & 61.09\ci{0.70}      & 75.32\ci{0.50}      & 74.24\ci{0.68}      & 84.09\ci{0.42}      \\ 
Transductive            & \ch & \ch  &      &        &     & 65.04\ci{0.75}      & 76.82\ci{0.50}      & 79.42\ci{0.69}      & 85.34\ci{0.43}      \\
Transductive            & \ch &      & \ch  &        &     & 65.57\ci{0.89}      & 78.03\ci{0.54}      & 78.29\ci{0.76}      & 88.02\ci{0.41}      \\ 
Transductive$\dagger$   & \ch & \ch  & \ch  &        &     & \tb{69.79}\ci{0.99} & 79.82\ci{0.55}      & \tb{83.05}\ci{0.79} & 88.82\ci{0.42} \\ 
Transductive            & \ch & \ch  &      & \ch    &     & 58.27\ci{0.91}      & 74.11\ci{0.56}      & 80.75\ci{0.76}      & 87.62\ci{0.44}      \\ 
Transductive            &     & \ch  & \ch  &        & \ch & 68.79\ci{0.96}      & \tb{79.93}\ci{0.56} & 82.04\ci{0.78}      & \tb{88.89}\ci{0.41}      \\
\bottomrule
\end{tabular}
\vspace{6pt}
\caption{\emph{Ablation study of algorithmic components} of our method \ours on \emph{mini}ImageNet. Inductive: baseline using only support examples. \lp: label propagation. \bal: class balancing~\eq{balance}. \ilc: iterative label cleaning, without which we just output predictions~\eq{argmax}. \iprob: iterative selection of top examples per class directly as column-wise maxima of $P$~\eq{sub} instead of \ilc. \cp: linear classifier used for prediction instead of \lp, as in~\cite{lrici}, with balancing still applied on output probabilities. $\dagger$: default setting of \ours.}
\label{tab:ab-comp}
\end{table*}

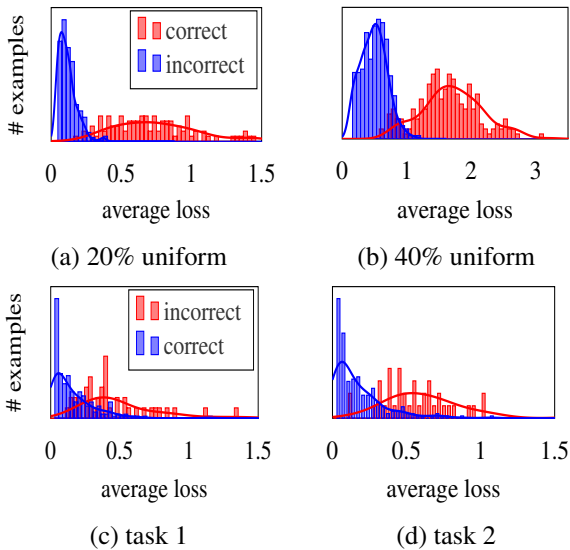
\begin{figure}
\begin{tabular}{cc}
\small
\centering
\setlength\tabcolsep{0pt}

\extfig{art-20}
{
\begin{tikzpicture}
\begin{axis}[
	width=.55\linewidth,
	height=.38\linewidth,
	font=\small,
	ybar,
	xmin=0,
	xmax=1.5,
	ymin=0,
	xlabel={average loss},
	ylabel={\# examples},
	yticklabels={,,},
	xtick style={draw=none},
	ytick style={draw=none},
	grid=none,
	bar/.style={thin,fill opacity=0.5,bar width=0.029497127532958984},
]
\addplot[red, fill=red, bar] table[x=bins,y=bad] \weightHistInit; \leg{incorrect}
\addplot[blue,fill=blue,bar] table[x=bins,y=ok]  \weightHistInit; \leg{correct}
\addplot[domain=0.2:1.5,smooth,red] table[x=x_noisy,y=kde_noisy] \labelkdea;
\addplot[domain=0:0.5,smooth,blue] table[x=x_clean,y=kde_clean] \labelkdea;
\end{axis}
\end{tikzpicture}
}

&

\extfig{art-40}
{
\begin{tikzpicture}
\begin{axis}[
    width=.55\linewidth,
    height=.38\linewidth,
    font=\small,
    ybar,
    xmax=3.5,
    xmin=0,
    ymin=0,
    xlabel={average loss},
    yticklabels={,,},
    xtick style={draw=none},
    ytick style={draw=none},
    grid=none,
    bar/.style={thin,fill opacity=0.5,bar width=0.062066144943237304},
]
\addplot[red, fill=red, bar] table[x=bins,y=bad] \weightHistMiddle;
\addplot[blue,fill=blue,bar] table[x=bins,y=ok]  \weightHistMiddle;
\addplot[domain=0:4,smooth,red] table[x=x_noisy,y=kde_noisy] \labelkdeb;
\addplot[domain=0:3,samples=200,smooth,blue] table[x=x_clean,y=kde_clean] \labelkdeb;

\end{axis}
\end{tikzpicture}
}

\\
(a) 20\% uniform &
(b) 40\% uniform \\

\extfig{query-1}
{
\begin{tikzpicture}
\begin{axis}[
	width=.55\linewidth,
	height=.38\linewidth,
	font=\small,
	ybar,
	xmin=0,
	xmax=1.5,
	ymin=0,
	xlabel={average loss},
	ylabel={\# examples},
	yticklabels={,,},
	xtick style={draw=none},
	ytick style={draw=none},
	grid=none,
	bar/.style={thin,fill opacity=0.5,bar width=0.02777163028717041},
]
\addplot[red, fill=red, bar] table[x=bins,y=bad] \labelquerya; 
\addplot[blue,fill=blue,bar] table[x=bins,y=ok]  \labelquerya; 
\addplot[domain=0.2:1.5,smooth,red] table[x=x_noisy,y=kde_noisy] \labelqueryakde;
\addplot[domain=0:0.5,smooth,blue] table[x=x_clean,y=kde_clean] \labelqueryakde;
\end{axis}
\end{tikzpicture}
}

&

\extfig{query-2}
{
\begin{tikzpicture}
\begin{axis}[
    width=.55\linewidth,
    height=.38\linewidth,
    font=\small,
    ybar,
    xmax=1.5,
    xmin=0,
    ymin=0,
    xlabel={average loss},
    yticklabels={,,},
    xtick style={draw=none},
    ytick style={draw=none},
    grid=none,
    bar/.style={thin,fill opacity=0.5,bar width=0.02024201154708862},
]
\addplot[red,fill=red,bar] table[x=bins,y=bad]  \labelqueryc;
\addplot[blue,fill=blue,bar] table[x=bins,y=ok]  \labelqueryc;
\addplot[domain=0:4,smooth,red] table[x=x_noisy,y=kde_noisy] \labelqueryckde;
\addplot[domain=0:3,samples=200,smooth,blue] table[x=x_clean,y=kde_clean] \labelqueryckde;

\end{axis}
\end{tikzpicture}
}

\\
(c) task 1 &
(d) task 2

\end{tabular}
\caption{(a,b) Distributions of loss values~\eq{loss} for correctly and incorrectly labeled examples, normalized independently. Uniform label noise: (a) 20\%, (b) 40\%. Pseudo-labels predicted by~\eq{argmax} for two different 1-shot transductive \emph{mini}ImageNet tasks (c,d).}
\label{fig:hist}
\end{figure}

\begin{table*}
\small
\centering
\begin{tabular}{lccccccccc}
\toprule
\Th{Method}             & \Th{Network} & \mc{2}{\Th{\emph{mini}ImageNet}}          & \mc{2}{\Th{\emph{tiered}ImageNet}}        & \mc{2}{\Th{Cifar-FS}}                     & \mc{2}{\Th{CUB}}                          \\ 
                        &              & 1-shot              & 5-shot              & 1-shot              & 5-shot              & 1-shot              & 5-shot              & 1-shot       & 5-shot                     \\ \midrule
LR+ICI~\cite{lrici}     & ResNet-12A   & 66.80\cip           & 79.26\cip           & 80.79\cip           & 87.92\cip           & 73.97\cip           & 84.13\cip           & 88.06\cip           & 92.53\cip           \\
LR+ICI~\cite{lrici}*    & ResNet-12A   & 66.85\ci{0.92}      & 78.89\ci{0.55}      & 82.40\ci{0.84}      & 88.80\ci{0.50}      & 75.36\ci{0.97}      & 84.57\ci{0.57}      & 86.53\ci{0.79}      & 92.11\ci{0.35}      \\
\tb{\ours (ours)}       & ResNet-12A   & \tb{69.79}\ci{0.99} & \tb{79.82}\ci{0.55} & \tb{83.49}\ci{0.88} & \tb{89.48}\ci{0.47} & \tb{77.14}\ci{0.95} & \tb{85.23}\ci{0.55} & \tb{89.00}\ci{0.70} & \tb{92.74}\ci{0.35} \\ \midrule
PT+MAP~\cite{pt_map}    & WRN-28-10    & 82.92\ci{0.26}      & 88.82\ci{0.13}      & -                   & -                   & 87.69\ci{0.23}      & 90.68\ci{0.15}      & 91.55\ci{0.19}      & 93.99\ci{0.10}      \\
PT+MAP~\cite{pt_map}*   & WRN-28-10    & 82.88\ci{0.73}      & 88.78\ci{0.40}      & 88.15\ci{0.71}      & 92.32\ci{0.40}      & \tb{86.91}\ci{0.72}      & 90.50\ci{0.49}      & \tb{91.37}\ci{0.61}      & 93.93\ci{0.32}      \\
LR+ICI~\cite{lrici}*    & WRN-28-10    & 80.61\ci{0.80}      & 87.93\ci{0.44}      & 86.79\ci{0.76}      & 91.73\ci{0.40}      & 84.88\ci{0.79}      &  89.75\ci{0.48}     & 90.18\ci{0.65}      & 93.35\ci{0.30}      \\
\tb{\ours (ours)}       & WRN-28-10    & \tb{83.05}\ci{0.79} & \tb{88.82}\ci{0.42} & \tb{88.50}\ci{0.75} & \tb{92.46}\ci{0.42} & 86.51\ci{0.75} & \tb{90.60}\ci{0.48} & 91.03\ci{0.63} & \tb{94.11}\ci{0.30}\\ \bottomrule
\end{tabular}
\vspace{6pt}
\caption{\emph{Transductive inference}, comparison with LR+ICI~\cite{lrici} and PT+MAP~\cite{pt_map}. *: our reproduction with official code on our datasets.}
\label{tab:trans-pre}
\end{table*}

\begin{table*}
\small
\centering
\begin{tabular}{lccccccc}
\toprule
\Th{Method}             & \Th{Network} & \mc{2}{\Th{\emph{mini}ImageNet}}          & \mc{2}{\Th{\emph{tiered}ImageNet}}          & \mc{2}{\Th{Cifar-FS}}                     \\ 
                        &              & 1-shot              & 5-shot              & 1-shot              & 5-shot                & 1-shot              & 5-shot              \\ \midrule
LR+ICI~\cite{lrici}*    & WRN-28-10    & 82.38\ci{0.86}      & 88.78\ci{0.39}      & 88.59\ci{0.74}      & 92.11\ci{0.39}        & 86.39\ci{0.79}      & 90.02\ci{0.49}      \\
PT+MAP~\cite{pt_map}*   & WRN-28-10    & 83.79\ci{0.71}      & 88.94\ci{0.33}      & 88.87\ci{0.64}      & 92.01\ci{0.36}        & 87.63\ci{0.66}      & 90.15\ci{0.46}      \\
\tb{\ours (ours)}       & WRN-28-10    & \tb{85.98}\ci{0.74} & \tb{90.54}\ci{0.31} & \tb{90.02}\ci{0.70} &\tb{92.94}\ci{0.37}    & \tb{88.54}\ci{0.68} & \tb{90.92}\ci{0.46} \\ \bottomrule
\end{tabular}
\vspace{6pt}
\caption{\emph{Transductive inference, 50 queries per class}. *: our reproduction with official code on our datasets.}
\label{tab:trans-50q}
\end{table*}

\begin{table}
\small
\centering
\setlength\tabcolsep{4pt}
\begin{tabular}{lccccc}
\toprule
\Th{Method}             & \Th{\emph{m}IN}     & \Th{\emph{t}IN}     & \Th{Cifar-FS}       & \Th{CUB}            \\ \midrule
LR+ICI~\cite{lrici}*    & 88.69\ci{0.38}      & 91.88\ci{0.41}      & 90.23\ci{0.45}      & 93.66\ci{0.28}      \\
PT+MAP~\cite{pt_map}*   & 89.97\ci{0.34}      & 93.33\ci{0.34}      & 91.30\ci{0.45}      & 94.24\ci{0.28}      \\
\tb{\ours (ours)}       & \tb{90.51}\ci{0.35} & \tb{93.61}\ci{0.38} & \tb{91.59}\ci{0.44} & \tb{94.75}\ci{0.26} \\ \bottomrule
\end{tabular}
\vspace{6pt}
\caption{\emph{Transductive 10-shot inference} using WRN-28-10. \emph{m}IN: \emph{mini}ImageNet. \emph{t}IN: \emph{tiered}ImageNet. *: our reproduction with official code on our datasets.}
\label{tab:trans-10shot}
\end{table}


\begin{table}
\small
\centering
\setlength\tabcolsep{2pt}
\begin{tabular}{lccccc}
\toprule
\Th{Method}             & \Th{Pre} & \mc{2}{\emph{mini}ImageNet}               & \mc{2}{\emph{tiered}ImageNet}             \\ 
                        &          & 1-shot              & 5-shot              & 1-shot              & 5-shot              \\ \midrule
PT+MAP~\cite{pt_map}*   &          & 48.57\ci{0.81}      & 75.67\ci{0.82}      & 49.67\ci{0.77}      & 88.32\ci{0.50}      \\ 
\tb{\ours (ours)}       &          & 78.89\ci{0.90}      & 86.80\ci{0.46}      & 86.52\ci{0.47}      & 91.07\ci{0.47}      \\ \midrule
PT+MAP~\cite{pt_map}*   & \ch      & 82.88\ci{0.73}      & 88.78\ci{0.40}      & 88.15\ci{0.71}      & 92.32\ci{0.40}      \\
\tb{\ours (ours)}       & \ch    & \tb{83.05}\ci{0.79} & \tb{88.82}\ci{0.42} & \tb{88.50}\ci{0.75} & \tb{92.46}\ci{0.42} \\
\bottomrule
\end{tabular}
\vspace{6pt}
\caption{\emph{Transductive inference, ablation over PT+MAP~\cite{pt_map} pre-processing}. \Th{PRE}: power transform, normalization, centering. *: our reproduction with official code on our datasets.}
\label{tab:ab-pt}
\end{table}

\begin{figure}
\extfig{pca}
{
\begin{tikzpicture}
\begin{axis}[
	width=\linewidth,
	height=0.5\linewidth,
	font=\small,
	ymin=46,
	xlabel={dimensions},
	ylabel={mean accuracy},
	legend pos={south west},
]
\addplot[blue] table[x=d,y=ours] \pcadims; \leg{\ours (ours)}
\addplot[red]  table[x=d,y=ici]  \pcadims; \leg{LR+ICI}
\end{axis}
\end{tikzpicture}
}
\caption{\emph{1-shot trasductive inference} on \emph{mini}ImageNet, ablation over LR+ICI~\cite{lrici} pre-processing: dimension reduction by PCA.}
\label{fig:pca}
\end{figure}
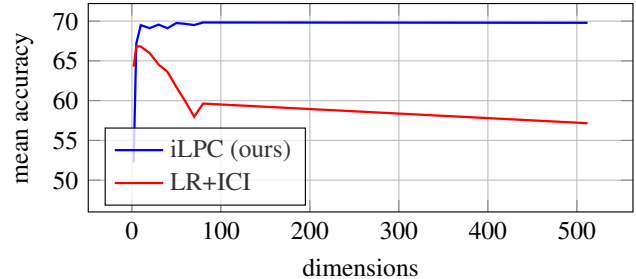

\begin{table}
\small
\centering
\setlength\tabcolsep{4pt}
\begin{tabular}{lcccccc}
\toprule
\Th{Method}                    & \Th{Netowrk} & \mc{2}{\Th{\emph{mini}ImageNet}}          \\  
                               &              & 1-shot              & 5-shot              \\ \midrule
MCT (instance,flip)~\cite{mct} & ResNet-12B   & 78.55\ci{0.86}      & 86.03\ci{0.42}      \\
MCT (no scale)~\cite{mct}*     & ResNet-12B   & 67.26\ci{0.60}      & \tb{81.90}\ci{0.43} \\
\tb{\ours (ours)}              & ResNet-12B   & \tb{75.58}\ci{1.16} & 81.58\ci{0.50}      \\ 
\ours (ours)                   & ResNet-12A   & 69.79\ci{0.99}      & 79.82\ci{0.55}      \\ \bottomrule
\end{tabular}
\vspace{6pt}
\caption{\emph{Transductive inference}, comparison with MCT~\cite{mct} using ResNet-12B. *: our reproduction with official code on our datasets, without augmentation and without scaling.}
\label{tab:trans-mct}
\end{table}

\begin{table*}
\small
\centering
\setlength{\tabcolsep}{6pt}
\begin{tabular}{lccccccccc}
\toprule
\Th{Balancing} & \Th{Network} & \mc{2}{\Th{\emph{mini}ImageNet}}          & \mc{2}{\Th{\emph{tiered}ImageNet}}        & \mc{2}{\Th{Cifar-FS}}                       & \mc{2}{\Th{CUB}}                          \\ 
               &              & 1-shot              & 5-shot              & 1-shot              & 5-shot              & 1-shot                & 5-shot              & 1-shot              & 5-shot              \\ \midrule
None           & WRN-28-10    & 78.06\ci{0.82}      & 87.80\ci{0.42}      & 86.04\ci{0.73}      & 90.74\ci{0.46}      & 85.32\ci{0.76}        &  89.64\ci{0.48}     & 89.67\ci{0.64}      & 92.98\ci{0.31}      \\
Uni            & WRN-28-10    & 77.50\ci{0.78}      & 83.68\ci{0.39}      & 83.02\ci{0.67}      & 86.17\ci{0.40}      & 81.47\ci{0.69}        & 84.83\ci{0.45}      & 85.22\ci{0.57}      & 87.99\ci{0.28}      \\
\tb{True}      & WRN-28-10    & \tb{82.68}\ci{0.82} & \tb{89.07}\ci{0.41} & \tb{89.17}\ci{0.70} & \tb{92.67}\ci{0.44} & \tb{87.32}\ci{0.74}   & \tb{90.92}\ci{0.48} & \tb{91.24}\ci{0.60} & \tb{94.14}\ci{0.30} \\ \bottomrule
\end{tabular}
\vspace{6pt}
\caption{\emph{Unbalanced transductive inference} with our \ours. Number of queries per class drawn uniformly at random from ${10,...,20}$. None: no balancing. Uni: Uniform distribution. True: True distribution.}
\label{tab:unbalanced}
\end{table*}



\subsection{Setup}
\label{sec:setup}
\paragraph{Datasets}

We use
four common few-shot classication benchmark datasets, \emph{mini}ImageNet~\cite{matchingNets, ravilstm}, \emph{tiered}ImageNet \cite{sslmeta}, CUB~\cite{closerlook, fewshotCUB} and CIFAR-FS~\cite{closerlook, cifar100db}.
More details are provided in the supplementary material.

\paragraph{Tasks}

We consider $N$-way, $K$-shot classification tasks with $N = 5$ randomly sampled novel classes and $K \in \{1, 5\}$ randomly selected examples per class for support set $S$, that is, $L = 5K$ examples in total. For the query set $Q$, we randomly sample $15$ additional examples per class, that is, $M = 75$ examples in total, which is the most common choice in the literature~\cite{tpn, learning2selftrain, transmatch}. 

In the semi-supervised setting, the unlabeled set $U$ contains an additional number of randomly sampled examples per novel class. This number depends on $K$. We use two settings, namely 30/50 and 100/100, where the first number (30 or 100) refers to 1-shot and the second (50 or 100) to 5-shot. Again, these are the two most common choices in semi-supervised few shot learning~\cite{learning2selftrain,lrici,mct,ssmeta,transmatch}.

Unless otherwise stated, we use 1000 tasks and report mean accuracy and 95\% confidence interval on the test set.

\paragraph{Competitors}

As discussed in the supplementary material, there are several flaws in experimental evaluation in the literature, like the use of different networks, training, versions of datasets, dimensionality and feature pre-processing. Fair comparison is impossible, unless one uses public code to reproduce results under exactly the same setup.

In this work, we do provide completely fair comparisons with such reproduced results of three state-of-the-art methods: LR+ICI~\cite{lrici}, PT+MAP~\cite{pt_map} and MCT~\cite{mct}. 
Only~\cite{lrici} is published, while the other two are pre-prints. 

\paragraph{Networks}

We use publicly available pre-trained backbone convolutional neural networks that are trained on the base-class training set. We experiment with two popular networks, namely, the residual network ResNet-12~\cite{tadam} and the wide residual network WRN-28-10~\cite{LEO}.

In particular, to compare with~\cite{lrici}, we use pre-trained weights of the ResNet-12 provided by~\cite{lrici}, which we call ResNet-12A, as well as official public code\footnote{\url{https://github.com/Yikai-Wang/ICI-FSL}} for testing. To compare with~\cite{pt_map}, we use pre-trained weights of a WRN-28-10 provided by~\cite{manifoldmixup}\footnote{\url{https://github.com/nupurkmr9/S2M2_fewshot}}, which are the same used by~\cite{pt_map}, as well as official public code\footnote{\url{https://github.com/yhu01/PT-MAP}} for testing. To compare with~\cite{mct}, we use official public code\footnote{\url{https://github.com/seongmin-kye/MCT}} to train from scratch another version of ResNet-12 used by~\cite{mct}, which we call ResNet-12B, as well as the same code for testing.

\paragraph{Feature pre-processing}

Each method uses its own feature pre-processing. LR+ICI~\cite{lrici} uses $\ell_2$-normalization and PCA to reduce ResNet-12A to 5 dimensions. PT+MAP~\cite{pt_map} uses element-wise power transform, $\ell_2$-normalization and centering of WRN-28-10 features. MCT~\cite{mct} uses flattening of the output tensor of ResNet-12B rather than spatial pooling.
By default, we use the same choices as~\cite{pt_map,mct} for WRN-28-10 and ResNet-12B. For ResNet-12A however, we use $\ell_2$-normalization only on transductive inference and we do not use any dimensionality reduction.

\paragraph{Implementation details}

We use PyTorch~\cite{pytorch}
and scikit-learn~\cite{sklearn}. Label cleaning is based on a linear classifier on top of $f$, initialized by imprinting the average of support features per class and then trained using~\eq{loss}. We use SGD with momentum 0.9 and weight decay 0.0005.  We use a learning rate of $\eta$ for 1000 iterations. For inductive (resp. semi-supervised) learning, we use logistic regression on support (resp. also pseudo-labeled) examples, learned using scikit-learn~\cite{lrici}. The row-wise sum $\vp$~\eq{sink-row} is fixed to 1. The supplementary material includes more choices.
It also includes inference time comparisons.


\subsection{Ablation study}
\label{sec:ablation}

\paragraph{Hyperparameters}

Our hyperparameters include $\gamma$ and $k$ used in the nearest neighbor graph~\eq{affinity}, $\alpha$ in label propagation~\eq{lp}, $\tau$ in balancing~\eq{balance} and the learning rate $\eta$ of label cleaning. We optimize them on the validation set of every dataset. Common choices for $k$ and $\alpha$ are in $[15,20]$ and in $[0.5,0.8]$, respectively. We set $\gamma = 3$, $\tau = 3$ and $\eta = 0.1$. Following~\cite{lrici}, we set $\nu = 3$ and select the 3 examples per class having the least average loss at every iteration in the transductive setting. We set $\nu = 1$ and $\nu = 5$ in the 1-shot and 5-shot settings respectively in the semi-supervised setting. More details and precise choices per dataset  are given in the supplementary material.

\paragraph{Algorithmic components}

\autoref{tab:ab-comp} ablates our method in the presence or not of individual components, as well as using alternative components. The use of queries with label propagation gives a gain of transductive over inductive inference, up to 6\% in 1-shot, while being on par with the linear classifier in 5-shot.
In 1-shot, balancing and iterative label cleaning bring another gain of 4-5\% each independently, while the combination of the two brings 8-9\%. The performance of iterative label cleaning is further justified by its superior of performance when compared to selecting examples based on $P$ instead.

\begin{table*}
\small
\centering
\setlength\tabcolsep{4.5pt}

\begin{tabular}{lcccccccccc}
\toprule
\Th{Method}                    & \Th{Network} & \Th{Split} & \mc{2}{\Th{\emph{mini}ImageNet}}          & \mc{2}{\Th{\emph{tiered}ImageNet}}        & \mc{2}{\Th{Cifar-FS}}                     & \mc{2}{\Th{CUB}}                 \\ 
                               &              &            & 1-shot              & 5-shot              & 1-shot              & 5-shot              & 1-shot              & 5-shot              & 1-shot               & 5-shot    \\ \midrule
LR+ICI~\cite{lrici}            & ResNet-12A   & 30/50      & 69.66\cip           & 80.11\cip           & 84.01\cip           & 89.00\cip           & 76.51\cip           & 84.32\cip           & 89.58\cip            & 92.48     \\
LR+ICI~\cite{lrici}*           & ResNet-12A   & 30/50      & 67.57\ci{0.97}      & 79.07\ci{0.56}      & 83.32\ci{0.87}      & 89.06\ci{0.51}      & 75.99\ci{0.98}      & 84.01\ci{0.62}      & 88.50\ci{0.71}       & -         \\

\tb{\ours (ours)}              & ResNet-12A   & 30/50      & \tb{70.99}\ci{0.91} & \tb{81.06}\ci{0.49} & \tb{85.04}\ci{0.79} & \tb{89.63}\ci{0.47} & \tb{78.57}\ci{0.80}      & \tb{85.84}\ci{0.56} & \tb{ 90.11}\ci{0.64} & -         \\ \midrule
LR+ICI~\cite{lrici}*         & WRN-28-10    & 30/50      & 81.31\ci{0.84}      & 88.53\ci{0.43}      & 88.48\ci{0.67}      & 92.03\ci{0.43}      & 86.03\ci{0.77}      & 89.57\ci{0.53}      & 90.82\ci{0.59}       & -         \\
PT+MAP~\cite{pt_map}$\dagger$ & WRN-28-10    & 30/50      & 83.14\ci{0.72}      & 88.95\ci{0.38}      & 89.16\ci{0.61}      & 92.30\ci{0.39}      & \tb{87.05}\ci{0.69}      & 89.98\ci{0.49}      & 91.52\ci{0.53}       & -         \\
\tb{\ours (ours)}             & WRN-28-10    & 30/50      & \tb{83.58}\ci{0.79} & \tb{89.68}\ci{0.37} & \tb{89.35}\ci{0.68} & \tb{92.61}\ci{0.39} & 87.03\ci{0.72} & \tb{90.34}\ci{0.50} & \tb{91.69}\ci{0.55}  & -         \\ 
\bottomrule
\end{tabular}
\vspace{6pt}
\caption{\emph{Semi-supervised few-shot learning}, comparison with~\cite{lrici, pt_map}. *: our reproduction with official code on our datasets. $\dagger$: our adaptation to semi-supervised, based on official code. CUB 5-shot omitted: no class has the required 70 examples.}
\label{tab:semi_pre}
\end{table*}
\begin{table*}
\small
\centering
\setlength\tabcolsep{5pt}
\begin{tabular}{lccccccccc}
\toprule
\Th{Method}                             & \Th{Network} & \mc{2}{\Th{\emph{mini}ImageNet}}          & \mc{2}{\Th{\emph{tiered}ImageNet}}        & \mc{2}{\Th{Cifar-FS}}                     & \mc{2}{\Th{CUB}}                          \\ 
                                        &              & 1-shot              & 5-shot              & 1-shot              & 5-shot              & 1-shot              & 5-shot              & 1-shot              & 5-shot              \\ \midrule
LR+ICI~\cite{lrici}*                    & ResNet-12A   & 66.85\ci{0.92}      & 78.89\ci{0.55}      & 82.40\ci{0.84}      & 88.80\ci{0.50}      & 75.36\ci{0.97}      & 84.57\ci{0.57}      & 86.53\ci{0.79}      & 92.11\ci{0.35}      \\
CAN+Top-\emph{k}~\cite{crossattention}  & ResNet-12    & 67.19\ci{0.55}      & 80.64\ci{0.35}      & 73.21\ci{0.58}      & 84.93 \ci{0.38}     & -                   & -                   & -                   & -                   \\
DPGN~\cite{DPGN}                        & ResNet-12    & 67.77\ci{0.32}      & 84.60\ci{0.43}      & 72.45\ci{0.51}      & 87.24\ci{0.39}      & 77.90\ci{0.50}      & 90.20\ci{0.40}      & 75.71\ci{0.47}      & 91.48\ci{0.33}      \\
MCT (instance)~\cite{mct}               & ResNet-12B   & 78.55\ci{0.86}      & 86.03\ci{0.42}      & 82.32\ci{0.81}      & 87.36\ci{0.50}      & 85.61\ci{0.69}      & 90.03\ci{0.46}      & -                   & -                   \\
\midrule
EP~\cite{embeddingpropagation}          & WRN-28-10    & 70.74\ci{0.85}      & 84.34\ci{0.53}      & 78.50\ci{0.91}      & 88.36\ci{0.57}      & -                   & -                   & -                   & -                   \\
SIB~\cite{SIB}$\dagger$                 & WRN-28-10    & 70.00\ci{0.60}      & 79.20\ci{0.40}      & 72.90\cip           & 82.80\cip           & 80.00\ci{0.60}      & 85.3\ci{0.40}       & -                   & -                   \\
SIB+E$^3$BM~\cite{ensemblefsl}          & WRN-28-10    & 71.40\ci{0.50}      & 81.20\ci{0.40}      & 75.60\ci{0.6}       & 84.30\ci{0.4}       & -                   & -                   & -                   & -                   \\
LaplacianShot~\cite{laplacianshot} & WRN-28-10 & 74.86\ci{0.19} & 84.13\ci{0.14} & 80.18\ci{0.21} & 87.56\ci{0.15} & - &- & -& -\\
PT+MAP~\cite{pt_map}*                   & WRN-28-10    & 82.88\ci{0.73}      & 88.78\ci{0.40}      & 88.15\ci{0.71}      & 92.32\ci{0.40}      & \tb{86.91}\ci{0.72}      & 90.50\ci{0.49}      & \tb{91.37}\ci{0.61}      & 93.93\ci{0.32}      \\
\tb{\ours (ours)}       & WRN-28-10    & \tb{83.05}\ci{0.79} & \tb{88.82}\ci{0.42} & \tb{88.50}\ci{0.75} & \tb{92.46}\ci{0.42} & 86.51\ci{0.75} & \tb{90.60}\ci{0.48} & 91.03\ci{0.63} & \tb{94.11}\ci{0.30}\\ \bottomrule
\end{tabular}
\vspace{6pt}
\caption{\emph{Transductive inference state of the art}. *: our reproduction with official code on our datasets. $\dagger$: \emph{tiered}ImageNet as reported by~\cite{ensemblefsl}.}
\label{tab:soa-trans}
\end{table*}

\begin{table}
\small
\centering
\setlength\tabcolsep{2pt}
\begin{tabular}{lcccccc}
\toprule
\Th{Method}                                        & \Th{Network} & \Th{Split} & \mc{2}{\Th{\emph{mini}ImageNet}}          \\ 
                                                   &              &            & 1-shot              & 5-shot              \\ \midrule
LST~\cite{learning2selftrain}                      & ResNet-12    & 30/50      & 70.10\ci{1.90}      & 78.70\ci{0.80}      \\
LR+ICI~\cite{lrici}                                & ResNet-12A   & 30/50      & 69.66\cip           & 80.11\cip           \\
MCT (instance)~\cite{mct}                          & ResNet-12B   & 30/50      & 73.80\ci{0.70}      & 84.40\ci{0.50}      \\
\midrule                                                                                            
$k$-means~\cite{ssmeta}$\dagger$                   & WRN-28-10    & 100/100    & 52.35\ci{0.89}      & 67.67\ci{0.65}      \\
TransMatch~\cite{transmatch}                       & WRN-28-10    & 100/100    & 63.02\ci{1.07}      & 81.06\ci{0.59}      \\
PTN~\cite{PTN} & WRN-28-10 & 100/100 &81.57\ci{0.94} & 87.17\ci{0.58} \\
\tb{\ours (ours)}                                  & WRN-28-10    & 100/100    & \tb{87.62}\ci{0.67}      & \tb{90.51}\ci{0.36} \\ \bottomrule
\end{tabular}
\vspace{6pt}
\caption{\emph{Semi-supervised few-shot learning state of the art}. $\dagger$: as reported by~\cite{transmatch}.}
\label{tab:soa-semi}
\end{table}

\subsection{Label cleaning: loss distribution}
\label{sec:hist}

To illustrate our label cleaning, we conduct two experiments, showing the distribution of the loss value~\eq{loss}. In the first, shown in \autoref{fig:hist}(a,b), we inject label noise uniformly at random to the 20\%~(a) and 40\%~(b) of 500 labeled examples. The correctly and incorrectly labeled examples have very different loss distributions. Importantly, while previous work on noisy labels~\cite{ArazoOAOM19,o2u,robustonfly} attempts to detect clean examples by an optimal threshold on the loss value, we only need few clean examples per iteration. Examples with minimal loss value are clean.

The second experiment is on two novel 1-shot transductive tasks, shown in \autoref{fig:hist}(c,d). We use 50 unlabeled queries per class and we predict pseudo-labels according to~\eq{argmax}. Label cleaning is more challenging here because the two distributions are more overlapping. This is natural because predictions are more informed than uniform, even if incorrect. Still, a large proportion of clean examples have a smaller loss value than the minimal value of noisy ones.


\subsection{Effectiveness of class balancing}

To show the effectiveness of our class balancing module, we carry out experiments in a novel setting for unbalanced few-shot transductive inference. In this setting, the number of queries per class for every few-shot task is drawn uniformly at random from $\{10\dots20\}$. We use no balancing, or we use balancing with uniform class distribution~\eq{balance}, or, assuming the prior class distribution $\vu \in \real^N$ is known, we replace $\frac{\vone_N}{N}$ in~\eq{balance} by $\vu$. As shown in \autoref{tab:unbalanced}, balancing improves accuracy by a large mangin, but only if the prior class distribution is known, otherwise it is harmful.


\subsection{Transductive inference}
\label{sec:trans}

\autoref{tab:trans-pre} compares our \ours with LR+ICI~\cite{lrici} and PT+MAP~\cite{pt_map} under the standard setting of 15 unlabelled queries per class. The truly fair comparison is with our reproductions, indicated by *. Apart from the default networks, we also use WRN-28-10 with LR+ICI~\cite{lrici}, since it is more powerful. Our \ours is on par with PT+MAP~\cite{pt_map} under this setting and superior to LR+ICI~\cite{lrici} by up to 3\% on \emph{mini}ImageNet 1-shot.

We also experiment with 50 unlabeled queries per class, or $M=250$ in total. As shown in \autoref{tab:trans-50q}, the gain over PT+MAP~\cite{pt_map} increases significantly, up to 2\% on \emph{mini}ImageNet 1-shot. This can be attributed to the fact that PT+MAP~\cite{pt_map} operates on Euclidean space, while we capture the manifold structure, which manifests itself in the presence of more data. A 10-shot experiment, 
is shown in \autoref{tab:trans-10shot}. The gain is around 0.5\%.

\autoref{tab:ab-pt} shows that PT+MAP~\cite{pt_map} is very sensitive to feature pre-processing, losing up to 40\% without it, while our \ours more robust, losing only up to 5\%. Similarly, \autoref{fig:pca} shows that LR+ICI~\cite{lrici} is sensitive to dimension reduction, working best at only 5 dimensions. By contrast, our \ours is very stable and only fails at 2 dimensions.

\autoref{tab:trans-mct} compares our \ours with MCT~\cite{mct}. We reproduce MCT results by training from scratch ResNet-12B using the official code and we test both methods without data augmentation (horizontal flipping) and without meta-learned scaling function. The objective is to compare the two transductive methods under the same backbone network and the same training process, which is clearly superior to ResNet-12A. Under these settings, MCT is slightly better in 5-shot but \ours outperforms it by a large margin in 1-shot.


\subsection{Semi-supervised learning}
\label{sec:semi}

As shown in \autoref{tab:semi_pre}, \ours is superior to LR+ICI~\cite{lrici} in all settings by an even larger margin than in transductive inference, \eg by nearly 3.5\% in \emph{mini}ImageNet 1-shot. This can be be attributed to capturing the manifold structure of the data, since there is more unlabeled data in this case. Because PT+MAP~\cite{pt_map} does not experiment with semi-supervised learning, we adapt it in the same way as ours, using the default WRN-28-10, outperforming it in most experiments.


\subsection{Comparison with the state of the art}
\label{sec:soa}

\autoref{tab:soa-trans} and \autoref{tab:soa-semi} compare our \ours with a larger collection of recent methods on the tranductive and semi-supervised settings, respectively. Even when the network and data split appears to be the same, we acknowledge that our results are not directly comparable with any method other than our reproductions. As discussed in the supplementary material, this is due to the very diverse choices made in the bibliography, \eg versions of network, training settings, versions of datasets, or pre-processing. For instance, ResNet-12 is different than either ResNet-12A or ResNet-12B.

For this reason, we focus on the best result by each method, including ours. Necessarily, methods experimenting with WRN-28-10 have an advantage. Still, at least among those, \ours performs best by a large margin in both settings, with the closest second best being PT+MAP~\cite{pt_map}.

\section{Conclusion}
\label{sec:conclusion}

Our solution is conceptually simple and combines in a unique way ideas that have been successful in problems related to our task at hand. Label propagation exploits the manifold structure of the data, which becomes important in the presence of more data, while still being competitive otherwise. Class balancing provides a strong hint in correcting predictions when certain classes dominate. Label cleaning, originally introduced for learning with noisy labels, is also very successful in cleaning predicted pseudo-labels. Iterative reuse of few pseudo-labels as true labels bypasses the difficulty of single-shot detection of clean examples. 

Importantly, reasonable baselines, like predicting pseudo-labels by a classifier or iteratively re-using pseudo-labels without cleaning, fail completely. When compared under fair settings, our \ours outperforms or is on par with state-of-the art methods.
It is also significantly more robust against feature pre-processing 
on which other methods rely. 

{\small
\bibliographystyle{ieee_fullname}
\bibliography{tex/references}
}

\clearpage





\appendix
\begin{center}
\textbf{\Large Supplementary material}
\end{center}
\section{Datasets}
\label{app:datasets}

\paragraph{\emph{mini}ImageNet}

This is a widely used few-shot image classification dataset~\cite{matchingNets, ravilstm}. It contains 100 randomly sampled classes from ImageNet~\cite{imagenet}. These 100 classes are split into 64 training (base) classes, 16 validation (novel) classes and 20 test (novel) classes. Each class contains 600 examples (images). We follow the commonly used split proposed in~\cite{ravilstm}. All images are resized to $84 \times 84$.

\paragraph{\emph{tiered}ImageNet}

This is also sampled from ImageNet~\cite{imagenet} but has a hierarchical structure. Classes are partitioned into 34 categories, organized into 20 training, 6 validation and 8 test categories, containing 351, 97 and 160 classes, respectively. This ensures that training classes are semantically distinct from test classes, which is more realistic. We follow the common split of~\cite{sslmeta}. Again, all images are $84 \times 84$.

\paragraph{CUB}

This is a fine-grained classification dataset consisting of 200 classes, each corresponding to a bird species. We follow the split defined by~\cite{closerlook, fewshotCUB}, with 100 training, 50 validation and 50 test classes. To compare fairly with competitors, we \emph{use} bounding boxes on ResNet features following~\cite{fewshotCUBbounding} to compare with~\cite{lrici} but we \emph{do not use} bounding boxes on WRN~\cite{LEO} features to compare with~\cite{pt_map}.

\paragraph{CIFAR-FS}

This dataset is derived from CIFAR-100~\cite{cifar100db}, consisting of 100 classes with 600 examples per class. We follow the split provided by~\cite{closerlook}, with 64 training, 16 validation and 20 test classes. To compare fairly with competitors, we use the original image resolution of $32 \times 32$ on WRN features to compare with~\cite{pt_map} but we resize images to $84 \times 84$ on ResNet features to compare with~\cite{lrici}.


\section{Feature pre-processing}
\label{sec:preprocessing}

\paragraph{ResNet-12A}

ResNet-12A is the pre-trained backbone network used in~\cite{lrici}. For all of our transductive and semi-supervised experiments using this network, we adopt exactly the same pre-processing as~\cite{lrici}, which is $\ell_2$-normalization on the output features.

\paragraph{WRN-28-10}

WRN-28-10 is the pre-trained network used in \cite{manifoldmixup} and \cite{pt_map}. To provide fair comparisons with PT+MAP \cite{pt_map} we adopt exactly the same pre-processing as~\cite{pt_map}. In the transductive experiments, we apply power transform, $\ell_2$-normalization and centering on the output features. In the semi-supervised experiments, we applied centering by calculating the mean and variance from the support set, $S$, and the unlabeled  set, $U$, not taking into consideration the query set, $Q$, since this is an inductive setting.

\paragraph{ResNet-12B}

ResNet-12B is the pre-trained network used in MCT~\cite{mct}. For the experiments in \autoref{tab:trans-mct}, we adopt exactly the same pre-processing as~\cite{mct}, that is, $\ell_2$-normalization on the output features.


\section{Hyperparameters}
\label{sec:hyper}

\autoref{tab:param} shows the best hyperparameters $k$~\eq{affinity} and $\alpha$~\eq{lp} for every dataset, network and number of support examples per class $K \in \{1, 5\}$. The hyperparameters are optimized on the validation set separately for each experiment. We carried out experiements in the transductive setting for $k \in \{5,8,10, 15, 20, 25, 30, 40,50, 60\}$ and $\alpha \in \{0.2, 0.3, 0.4, 0.5, 0.6, 0.7, 0.8, 0.9\}$ and select the combination resulting in the best mean validation accuracy. In the semi-supervised setting, we use the same optimal values.


\section{Confidence weights}
\label{sec:weights}

The Sinkhorn-Knopp algorithm iteratively normalizes a $M \times N$ positive matrix $P$ to a row-wise sum $\vp \in \real^M$ and column-wise sum $\vq \in \real^N$. We experiment with two ways of setting the value of $\vp$:

\begin{enumerate}

	\item \textbf{Uniform.} Interpreting the $i$-th row of $P$ as a class probability distribution for the $i$-th query, it should be normalized to one, such that $p_i = 1$ uniformly.

	\item \textbf{Entropy.} Because we do not have the same confidence for each prediction, we use the entropy of the predicted class probability distribution of each example to quantify its uncertainty. Following~\cite{semilpavrithis}, we associate to each example $x_{L+i}$ for $i \in [M]$ a weight
	\begin{equation}
		\omega_i \defn 1 - \frac{H(\mathbf{\hat{z_i}})}{\log{(N)}},
	\label{eq:entropy}
	\end{equation}
	where $N$ is the number of classes and $\hat{z}_i$ is the $\ell_1$-normalized $i$-th row of $Z$~\eq{lp}, that is, $\hat{z}_{ij} \defn z_{ij}/\sum_{k = 1}^Nz_{ik}$. We then set the confidence weights $p_i = \omega_i$. Note that $\omega_i$ takes values in $[0, 1]$ because $\log(N)$ is the maximum possible entropy.
\end{enumerate}

Given $\vp$ and assuming balanced classes, $\vq$ is defined by~\eq{balance}, that is, $q_j =\frac{1}{N} \sum_{i=1}^Mp_i$ for $j \in [N]$. In the special case of $p_i = 1$, this simplifies to $q_j = \frac{M}{N}$.

\autoref{tab:sink-comparison} compares the two approaches. Even though using non-uniform confidence weights is a reasonable choice, uniform weights are superior in all settings. This can be attributed to the fact that examples with small weight tend to be ignored in the balancing process, hence their class distribution and consequently their predictions are determinded mostly by other examples with large weight. For this reason, examples with small weight may get more incorrect predictions in the case of entropy.


\begin{table}
\small
\centering
\begin{tabular}{lcccccccc}
\toprule
\Th{Param} & \mc{2}{\Th{\emph{m}IN}} & \mc{2}{\Th{\emph{t}IN}} & \mc{2}{\Th{CFS}} & \mc{2}{\Th{CUB}} \\ \midrule
$K$ (shot)        & 1   & 5   & 1   & 5   & 1   & 5   & 1   & 5   \\ \midrule
\mc{9}{ResNet-12A}                                                \\ \midrule
$k$~\eq{affinity} & 15  & 25  & 15  & 60  & 15  & 15  & 10  & 8   \\
$\alpha$~\eq{lp}  & 0.8 & 0.4 & 0.5 & 0.8 & 0.8 & 0.4 & 0.6 & 0.6 \\ \midrule
\mc{9}{ResNet-12B}                                                \\ \midrule
$k$~\eq{affinity} & 15  & 15  & -   & -   & -   & -   & -   & -   \\
$\alpha$~\eq{lp}  & 0.9 & 0.9 & -   & -   & -   & -   & -   & -   \\ \midrule
\mc{9}{WRN-28-10}                                                 \\ \midrule
$k$~\eq{affinity} & 20  & 30  & 20  & 20  & 20  & 25  & 25  & 25  \\
$\alpha$~\eq{lp}  & 0.8 & 0.2 & 0.8 & 0.8 & 0.4 & 0.5 & 0.2 & 0.5 \\ \bottomrule
\end{tabular}
\vspace{6pt}
\caption{Selected hyperparameters. \emph{m}IN: \emph{mini}ImageNet. \emph{t}IN: \emph{tiered}ImageNet. CFS: CIFAR-FS.}
\label{tab:param}
\end{table}

\begin{table}
\small
\centering
\setlength\tabcolsep{2pt}
\begin{tabular}{lccccc}
\toprule
\Th{Method}                    & \mc{2}{\Th{ResNet-12A}}                    & \mc{2}{\Th{WRN-28-10}}                    \\
                               & 1-shot              & 5-shot              & 1-shot              & 5-shot              \\ \midrule
uniform                        & \tb{69.79}\ci{0.99} & \tb{79.82}\ci{0.55} & \tb{83.05}\ci{0.79} & \tb{88.82}\ci{0.42} \\
entropy                        & 66.94\ci{1.01}      & 78.34\ci{0.58}      & 81.05\ci{0.90}      & 88.43\ci{0.44}      \\
\bottomrule
\end{tabular}
\vspace{6pt}
\caption{Comparison between ways of setting confidence weights $\vp$; transductive inference on \emph{mini}ImageNet. Uniform: $p_i = 1$. Entropy: $p_i = \omega_i$~\eq{entropy}.}
\label{tab:sink-comparison}
\end{table}

\section{Inference time}

We conduct inference time experiments to investigate the computational efficiency of our \ours compared with PT+MAP~\cite{pt_map} and LR+ICI~\cite{lrici}. Using the WRN-28-10 backbone, we calculate the inference time required for a single 5-way, 1-shot task, averaged over 1000 tasks. For each task there are 15 queries per class. The results can be seen on \autoref{tab:inference_times}.

\begin{table}
\small
\centering
\setlength\tabcolsep{2pt}
\begin{tabular}{lc}
\toprule
\Th{Method}    & {\Th{Inferene time}}                    \\ \midrule
LR+ICI~\cite{lrici}  & 0.89\\
\tb{PT+MAP~\cite{pt_map}}  &   \tb{0.57} \\
\ours & 1.20 \\
\bottomrule
\end{tabular}
\vspace{6pt}
\caption{Average inference time (in seconds) for the 1-shot
tasks in \emph{mini}ImageNet dataset.}
\label{tab:inference_times}
\end{table}

\section{Flaws in evaluation}
\label{sec:flaws}

Throughout our investigations we observed that comparisons are commonly published that are not under the same settings. In this section we highlight such problems.

\begin{enumerate}
    \item In multiple works such as~\cite{embeddingpropagation, pt_map, DPGN, baseline_ft}, comparisons between state-of-the-art methods are made without explicitly differentiating between inductive and transductive methods. This is unfair since transductive methods perform better by leveraging query data.

    \item Comparisons use different networks without mentioning so. For example, Table 1 of~\cite{transmatch} does not indicate what network each method uses. \cite{transmatch} uses WRN-28-10, while~\cite{tpn} uses a 4-layer convolutional network.

    \item In the semi-supervised setting, comparisons use different numbers of unlabelled data without mentioning so. In Table 4 of~\cite{embeddingpropagation} for example, \cite{embeddingpropagation} uses 100 unlabelled examples while~\cite{learning2selftrain} uses 30 for 1-shot and 50 for 5-shot, \cite{ssmeta} and \cite{tpn} use 20 for 1-shot and 20 for 5-shot. In Table 1 of~\cite{lrici}, the best model of~\cite{lrici} uses an 80/80 split for 1/5-shot while other methods such as~\cite{learning2selftrain} use a 30/50 split. In Table 1 of~\cite{transmatch}, ~\cite{transmatch} uses 100 or 200 unlabelled examples while~\cite{ssmeta,tpn} use 20/20 split for 1/5-shot.

    \item Some methods use different dataset settings when comparing with other methods without explicitly stating so. In Table 1 of~\cite{lrici} for instance, \cite{lrici} uses the bounding box provided for CUB while other methods such as \cite{closerlook,metaoptnet} do not.

    \item Comparisons using the same network is made but this network has been trained using a different training regimes. Unless the novelty of the work lies in the training regime, this is unfair. As shown in~\cite{manifoldmixup}, a better training regime can increase the performance significantly.

    \item There are several different variants of the benchmark datasets, coming from different sources. The two most common variants are~\cite{closerlook}, which uses original image files, and~\cite{metaoptnet}, which uses pre-processed tensors stored in \texttt{pkl} files. Testing a network on a different variant than the one it was trained on may result in performance drops as large as 5\%.
\end{enumerate}

We believe that highlighting these evaluation flaws will help researchers avoid making such mistakes and move towards a fairer evaluation. We encourage the community to compare different methods against the same settings and if otherwise, state clearly the differences. As a contribution towards a fairer evaluation, we intend to make our code publicly available along with the pre-trained networks used in this work.


\end{document}